\documentclass[a4paper]{article}
\usepackage{tabularx}
\usepackage{paralist}
\usepackage{INTERSPEECH2022}

\title{Improving Proactive Dialog Agents Using Socially-Aware Reinforcement Learning (Preprint)}
\name{Matthias Kraus$^1$, Nicolas Wagner$^2$, Ron Riekenbrauck$^2$,Wolfgang Minker$^2$}
\address{
  $^1$Augsburg University\\
  $^2$Ulm University}
\email{matthias.kraus@uni-a.de, nicolas.wagner@uni-ulm.de, ron.riekenbrauck@uni-ulm.de, wolfgang-minker@uni-ulm.de}

\begin{document}

\maketitle
\begin{abstract}
  The next step for intelligent dialog agents is to escape their role as silent bystanders and become proactive. Well-defined proactive behavior may improve human-machine cooperation, as the agent takes a more active role during interaction and takes off responsibility from the user. However, proactivity is a double-edged sword because poorly executed pre-emptive actions may have a devastating effect not only on the task outcome but also on the relationship with the user. For designing adequate proactive dialog strategies, we propose a novel approach including both socially as well as task-relevant features in the dialog. Here, the primary goal is to optimize proactive behavior so that it is task-oriented -  this implies high task success and efficiency -  while also being socially effective by fostering user trust. Including both aspects in the reward function for training a proactive dialog agent using reinforcement learning showed the benefit of our approach for a more successful human-machine cooperation.

\end{abstract}

\noindent\textbf{Preprint of the following publication:}:\textit{ Matthias Kraus, Nicolas Wagner, Ron Riekenbrauck, and Wolfgang Minker. 2023. Improving Proactive Dialog Agents Using Socially-Aware Reinforcement Learning. In Proceedings of the 31st ACM Conference on User Modeling, Adaptation and Personalization (UMAP’23), June 26–29, 2023, Limassol, Cyprus. ACM, New York, NY, USA, 16 pages. https://doi.org/10.1145/3565472.3595611}

\section{Introduction}
Conversational AI in the form of dialogue agents (DA) and chatbots are nowadays commonly applied in various areas helping with a diverse set of tasks. For example, Amazon Alexa enables people to manage their grocery shopping using natural language, whereas chatbots are used for digital customer-service. Although such agents are constantly evolving and are often described as intelligent, humans still perceive them in the role of a butler \cite{mcmillan2021leaving}. Thus, they are still only trusted fulfilling simple, less risky tasks, and are not applied for tasks that require a higher degree of cooperation, e.g. assisting in stock exchanges, or as a business advisor. 

A reason for this lack of trust stems from the low degree of conversational intelligence of contemporary systems. Chaves and Gerosa \cite{chaves2021should} describe a chatbot's conversational intelligence as ``to actively participate in the conversation and to demonstrate awareness of the topic discussed, the evolving conversational context, and the flow of the dialog''. However, current systems still mostly act in a reactive manner, i.e. they just react on commands, or express low-level proactive behavior for simple tasks \cite{sarikaya2017technology}, e.g. reminding the user of a specific event which the user explicitly set in the calendar or providing movie recommendations based on user habits or routines. 
For being trusted and effective in assisting with more complex tasks, the conversational intelligence of DAs needs to be improved. For achieving this, we deem sophisticated and sound proactive dialog strategies inevitable. Proactive behavior of conversational systems can be defined as self-initiated and anticipatory behavior \cite{nothdurft2015finding}. Providing systems with appropriate proactivity during cooperative tasks has shown to benefit a system's helpfulness \cite{peng2019design}, and user satisfaction \cite{baraglia2016initiative}. However, proactivity is a double-edged sword, where irrelevant or untimely interventions may have highly negative consequences. The probably best example in this regard forms Microsoft's infamous office assistant \textsc{Clippit}~\cite{horvitz1998lumiere}. 
The agent did not act according to the user's expectations and was perceived as distractive and not trustworthy, which resulted in a negative system reception \cite{bickmore2005establishing}. Therefore, the design, modelling, and implementation of effective and trusted proactive dialog is a delicate task.

We tackle this issue by presenting the first principled approach to proactive dialog modelling by extending the Reinforcement Learning (RL) based dialog management approach. In modular task-oriented dialog systems, the dialog management component is responsible for deciding on appropriate system responses taking into account a semantic representation of the user's input, as well as the dialog history, and other contextual information with the aim of fulfilling a specific user goal (e.g. see McTear \cite{mctear2020conversational}). For reactive assistance, RL-based dialog management approaches have shown to provide optimal results considering task success and efficiency \cite{young2013pomdp,lemon2008adaptive, rieser2011reinforcement}, as well as user satisfaction \cite{ultes2015quality,ultes2019improving}. Therefore, we adopt a RL approach for optimizing proactive dialog strategies with the goal of improving the conversational intelligence of a system and thus enhancing the human-machine cooperation. Unlike user-initiated reactive dialog systems, that usually focus on optimising solely task-related metrics, proactive dialog system also require a certain social awareness for acting in favour of the user expectations. Recent research \cite{kraus2022design} provides evidence that the level of perceived trust in the system reflects whether a proactive system acts in accordance with the user's expectations. Therefore, we trained a proactive DA in a simulated environment, in which the agent assisted the user with a sequential decision-making task, using both trust estimate and usability measures for providing adequate behavior. In a standardized user simulation evaluation, we compared the RL-based proactive DA with agents using only static and rule-based proactive dialog strategies. The results demonstrate that including trust in the dialog model (and in the reward function) for enabling trust-adaptive dialog proved to be successful for creating socially (i.e. ``human-likeness'') and task efficient agents, where they were to achieve the best compromise of contributing to task completion effectively, but also acting in a trustworthy manner. 
\section{Related Work}
\label{sec:rel}
\subsection{Proactive Conversational AI}
Proactivity has been an extensive research field both in human-computer interaction (HCI) and AI for the last decades \cite{meurisch2020exploring, yorke2012design, chaves2021should, horvitz1999principles, sarikaya2017technology}. However, the definitions for the concept of proactivity differ greatly between the individual research areas. In recommendation systems \cite{christakopoulou2016towards}, for example, proactive behavior is understood as suggesting specific items for simplifying a user's navigation in large product or information spaces. In non-task-oriented open-domain dialog systems, proactivity is understood as actively leading the dialog and changing the topic from a start to a goal topic (e.g. see Wu et al.; Zhu et al.; Tang et al.; Xu et al. \cite{wu2019proactive,zhu2021proactive, tang2019target, xu2020knowledge}). In both areas, proactive conversation is often modelled using various combinations of knowledge graphs and deep learning models for providing personalized suggestions (recommender systems) or achieving a natural, coherent and engaging dialog flow (open-domain dialog). In personal intelligent assistants \cite{sarikaya2017technology, yorke2012design}, proactive behavior is closely related to concept of mixed-initiative interaction \cite{horvitz1999principles} and may be realized in multiple different ways. In such interactions, a user and an autonomous agent capable of acting on its own collaborate to solve tasks. To provide support, the agent must keep track of the user's activities and goals and weigh the costs and benefits of automated actions. Here, task-oriented proactive dialog is used to communicate and negotiate a system's decision-making process to minimize the risk of system failure and to solve task efficiently. We understand proactive dialog under this definition in the scope of this paper. Here, the essential challenges are the timing and the level of proactive actions.  Usually, the degree of proactive behavior in personal assistants can be modelled using different levels of autonomy (LoA). In this regard, Isbell and Pierce \cite{isbell2005ip} adopted the LoA from Sheridan and Verplank \cite{sheridan1978human} for defining the Interface-Proactivity (IP) continuum. The IP continuum consists of five levels and ranges from reactive behavior (``users do it themselves'') to completely autonomous system behavior (``system does it by itself''). Typically, proactive behavior of intelligent personal assistants may span several levels on that continuum. Due to the complexity of the task domains, usually rule-based approaches have been applied to guarantee reliable and predictable behavior. 

For example, \textsc{Radar} was a personal assistant that could help office employees to solve their tasks more efficiently \cite{faulring2010agent, garlan2007radar}. Similarly, \textsc{Calo} \cite{yorke2012design} was a proactive assistant that helped users with task management in an office environment. It could assist with organizing meetings or reminding of important activities. For deciding the level and timing of proactive dialog, e.g. whether a meeting was scheduled automatically or an interaction was initiated, both examples made use of task-specific dialog managers. They contained knowledge about a specific user's attention and interruption policies. Based on this information, the system could weigh the cost-benefit ratio of proactive behavior and adjust its level of proactivity accordingly. For example, if a high workload was detected, the system could suggest automatically preparing background material for the meeting or offering a reminder. The cost-benefit was calculated using system-related metrics such as the urgency of a proactive behavior or the cost of an error or interruption. The decision on which level of autonomy to choose was based on for the specific task, and fixed rules. Thus, the generated rules for proactive behavior may be not transferable to different scenarios and task domains. Further, they are quite costly to develop, due to the large set of necessary rules for reproducing adequate behavior. In addition, the rules were only designed with a limited set of users. Hence, they may not meet the expectations of all users and could be perceived as inflexible. For these reasons, we deem statistical methods for proactive dialog management beneficial. Especially, the RL paradigm seems to be predestined for proactive dialog management due to its nature of long-term planning and optimal decision-making under uncertainty. Therefore, we apply RL for deciding on optimal proactive system actions during mixed-initiative cooperation. For modelling proactive system behavior, we use several proactive dialog actions defined in previous work \cite{kraus2020effects}. They reflect the degree of interference of autonomous systems in human actions and will be explained in more detail in the Section 3. Contrary to reactive RL-based dialog management, it may not be beneficial to focus only on rewarding task-efficient behavior but also to include a certain social awareness. Due to a shift of control towards the agent. a potential loss of self-autonomy may occur. Therefore, a formation of trust for the user is required, otherwise the assistance possibly will be rejected and becomes obsolete \cite{schaefer2016meta}. For this reason, including a trust measure in the reward function needs to be considered. We shortly describe the concept of human-computer trust (HCT) in the following.   
\subsection{Human-Computer Trust}
Among the various definitions of trust in relationships between artificial agents and humans, we utilize the notion provided by Lee and See \cite{lee2004trust} who describe trust as ``the attitude that an agent will help achieve an individual's goal in a situation characterized by uncertainty and vulnerability''. This often used definition was deemed to be the most suitable for explaining the trust relationship during cooperation.
According to Lee and See \cite{lee2004trust}, three factors are proposed to be considered when modeling trust: the human, the autonomous partner, and the environment. Each factor has specific properties that influence the HCT relationship, e.g., gender of the user/agent, personality of the user/agent, degree of automation, anthropomorphism of the agent, type/difficulty of the task. For a more detailed information on the impact of the individual factors, we refer the reader to Schaefer et al. \cite{schaefer2016meta}. In this work, we utilize trust-related properties as features for estimating the user's trust in the proactive DA during cooperation. Details of the trust estimation module are described in Section 3.

In human-computer studies, the user's trust in a system is usually measured using subjective measurements based on self-reported questionnaires. For example, Madsen and Gregor \cite{madsen2000measuring} developed a questionnaire based on a hierarchical model of trust where subjects can agree or disagree with statements about the system's trustworthiness. Here, the authors differentiate between affect-based and cognition-based trust. While affect-based trust, mostly refers to attributes relevant for a long-term relationships, cognition-based trust encompasses features relevant for short-term interaction, such as perceived understandability, perceived technical competence, and perceived reliability. In short-term interactions, mostly the functionality and usability of a system are of importance. Especially the system's competence and reliability are proven to have a large impact on the HCT relationship~\cite{muir1996trust,lee1994trust}.
In the scope of this paper, we refer to trust by means of cognition-based trust as we observe rather short-term interactions with the proactive DA.
For designing conversational strategies to achieve both task success and socially effective interaction between user and system, \cite{pecune2020framework} studied an RL-based approach. For this, the authors made use of a user simulator that included a rapport estimation module \cite{jain2018user} and equipped a conversational system with task-oriented and rapport building behavior, e.g. small talk, self-disclosure. Further, they included the estimated rapport besides task metrics in the reward function for optimising both task and social dialog policies. Training and testing the agent with the social user simulator showed the usefulness of their approach.

In summary, we study how to include the concept of trust for RL-based proactive dialog management. For this, we contribute a simulation-based framework for equipping proactive DAs with social awareness concerning the HCT relationship between user and agent. For integrating socially awareness into conversational systems, Pecune and Marsella \cite{pecune2020framework} also studied an RL-based approach. Contrary to our work, the authors made use of a user simulator that included a rapport estimation module \cite{jain2018user} and equipped a conversational system with task-oriented and rapport building behavior, e.g. small talk, self-disclosure. Further, they included the estimated rapport besides task metrics in the reward function for optimising both task and social dialog policies. Training and testing the agent with the social user simulator showed the usefulness of their approach. In this paper, we contribute an evaluation of our proposed approach for testing whether a socially-aware proactive DA enhances cooperation by measuring task success, duration and the user's trust in the agent's actions.

\label{sec:sim}
\begin{figure*} 
\centering
	\includegraphics[scale=0.5]{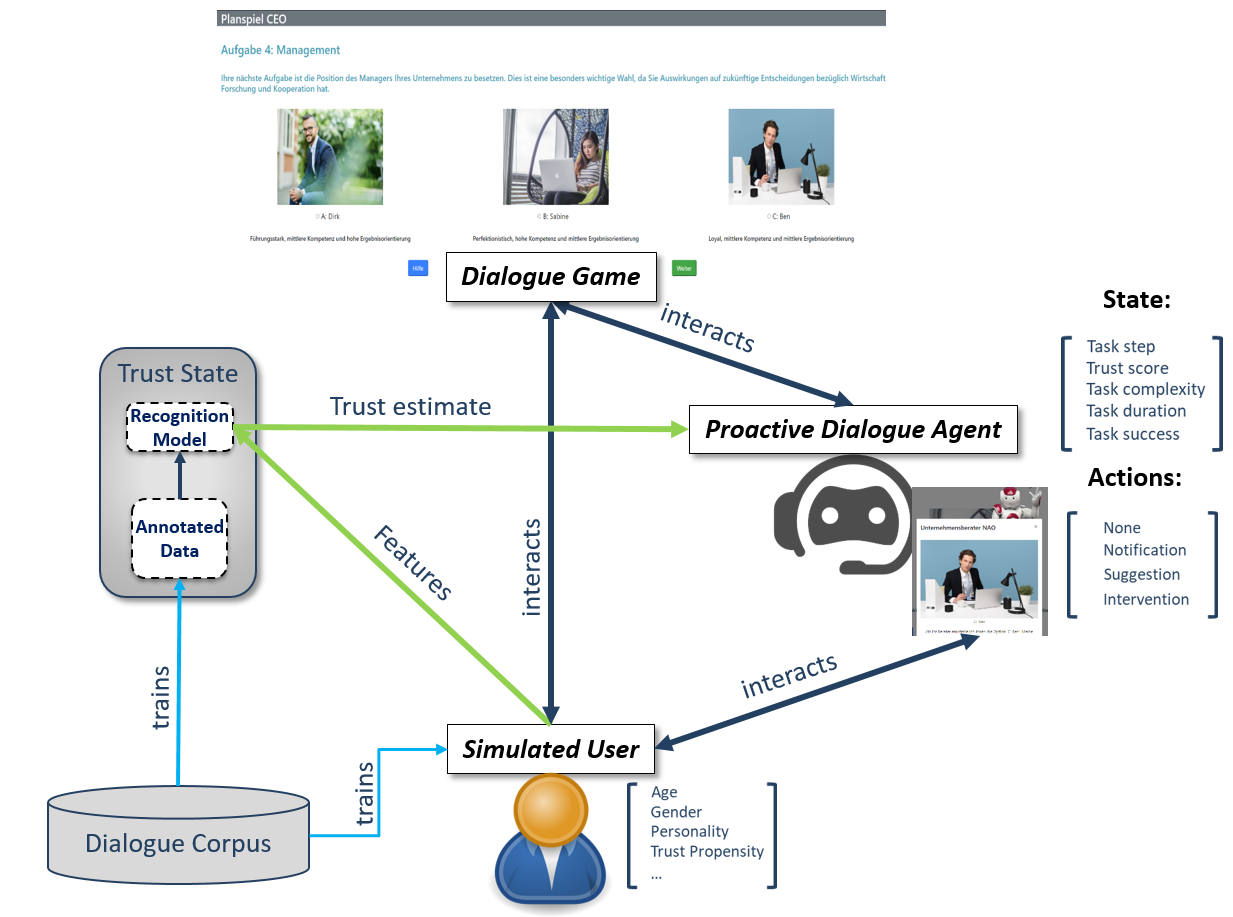}
	\caption{Components and information flows of the simulated RL-based proactive dialog environment.}
	\label{img:arch}
\end{figure*}
Typically, reactive RL-based DAs are trained using some kind of bootstrapping approach \cite{rieser2011reinforcement}. According to this approach, firstly, dialog data is collected using a Wizard-of-Oz (WoZ) approach. Based on this data set, different components of the simulated dialog environment can be build, e.g. user simulation, or noise model. Dialog policies can then be trained and evaluated in interaction with this simulated environment.

For training proactive dialog policies with an RL-based DA, we used a similar approach. However, data was not collected in a WoZ fashion but with a simplistic proactive DA. Moreover, for enabling socially-aware RL, we integrated a trust estimation module in the simulated environment for measuring the simulated user's trust level. The trust estimate was included both in the state space and the reward function of the RL-based proactive DA. In the following, we describe the bootstrap approach: First, we shortly review the data collection process based on previous work, and then detail the novel components that we developed for training a socially-aware RL-based proactive DA in a simulated environment. A depiction of the environment and interaction flows is depicted in Fig. \ref{img:arch}.
\section{Collecting Proactive Dialog Data for Socially-Aware RL-based Dialog Management}
In previous work \cite{kraus2022prodial}, a proactive dialog corpus consisting of 308 dialogs (3696 system-user exchanges) was created. For this, 308 users participated in an online data collection via the clickworker\footnote{www.clickworker.de} framework. During the data collection, users had to cooperate with a proactive DA in a serious dialog game, which was modelled as a sequential decision-making problem. The game consisted of 12 task steps, which were noted as system-user exchanges. The main difference between proactive dialog systems assisting with decision-making in comparison to reactive information-seeking systems is that the dialog goal is both known to the agent and the user. Thus, its task is not to identify the user's intention but to provide timely and appropriate assistance during task execution. 
Each exchange was annotated with self-reported measures on the system's trustworthiness and its related concepts user's perceived competence, predictability, reliability to represent the user's cognition-based trust. These variables were labeled on a 5-point Likert scale and ranged from 1="very low" to 5="very high". Further, we captured objective features such as task-related properties (complexity, exchange duration, user actions) as well as the system's actions, and static user information provided in a pre-test questionnaire, e.g. age, gender, personality, domain expertise. This allowed to developed a model for estimating the user's trust state for each exchange during the dialog game. In the following, we describe the main components of the data collection.
\subsection{Dialog Game}
The dialog game was implemented as a servlet-application based on a client-server model. On the client-side a user played the game and interacted with the proactive assistant using a clickable graphical user interface (GUI). On the server-side a self-implemented dialog control logic received user input from the GUI and provided task-related content to the interface by accessing a database. The user's goal was to successfully manage the company through skilful strategic action and to maximize profits. In doing so, a user had to make step-by-step decisions and plan undertakings in the interest of the company, such as location planning or personnel management. Individual decisions had consequences and affected the success of the management. The game was designed as a turn-based planning task, in which the system sequentially presented a task step and the available choices. The user could take different actions and cooperatively solve the task with the DA. Thus, the structure of the game can be describes as follows: the system takes an action providing task step relevant information, upon the user takes an action solving the respective task step after which the cycle is repeated until the game ends after a total of 12 task steps. The order of the tasks was fixed and could not be changed by the user. For each step, several options were presented from which the user had to choose. The number of options changed from task to task and ranged from a minimum of three to a maximum of five options.
At each task step, the user could perform four actions: select an option, ask for help, explicitly ask the DA for a suggestion, and continue with the game. When asking for help, general information about the game is given, such as what previous decisions need to be considered in the current task step. 
To add cause and effect to the user's decision making, options were linked to numerical scores. This allowed previous decisions to directly influence the value of future actions. The concept of a game state was to create a vulnerable, yet engaging environment for the user. Performance mattered, which should lead to the need to trust the assistant.

For illustration, consider the following example: User Alice is currently required to make a decision on task ``Research'', where a plausible research direction with regard to the built up company needs to be chosen. This task is influenced by Alice selections in previous tasks ``Management'' and ``Banking''. Depending on the combination of selections in respective tasks, one of the four options (Hydrogen Drive, Autonomous Driving, Battery Research, Climate Neutral Production) would yield the most points, whereas in the worst case scenario a user would yield zero points. The game score was based on an artificial scoring model, particularly developed for this application.
\subsection{Proactive Dialog Agent}
An agent system was able to choose from four different dialog actions to serve as a personal advisor to the user. The agent used natural language based on dynamically created templates. With regard to data collection purposes, it was designed as an expert system to avoid unintended side effects of incompetent system behavior on its trustworthiness. This allowed only the effects of proactive levels on HCT to be considered. To select the best option per task step, the agent used a simple reasoning mechanism by having knowledge about previous user decisions and accordingly querying the game's evaluation model.

The proactive assistance was modeled according to the proactive dialog actions defined by Kraus et al. \cite{kraus2021role}: \textit{None}, \textit{Notification}, \textit{Suggestion}, and \textit{Intervention}. These actions are tiered and range from no intervention (non-intrusive) to full intervention (very intrusive). Since the scenario was a planning task and the user was asked to select what they thought was the best option for each task step, the purpose of the system behavior was to use natural language to provide helpful information and suggestions for the selection process. With the reactive \textit{None} action, the system waited for the user to explicitly ask for suggestions. The more conservative proactive actions, \textit{Notification} and \textit{Suggestion}, let the user confirm the wizard's suggestions and differ only in the degree of directness. The \textit{Intervention} actions took the responsibility completely out of the user's hands and autonomously chose an option.
Assisting with sequential decision-making, proactive actions can be considered as the initiation of sub-dialogs, where the assistant influences a user action. For data collection purposes, the agent utilized a combined rule-based and randomized strategy for creating a diverse data set. In the following a typical example dialog is presented (note, that the user was only able to select pre-defined natural language templates):
\begin{compactitem}
\item[]\textbf{(Task:``Management''; \#Options = 3; Agent's Action: Notification)}
\item[\textbf{Agent (A):}] I have a suggestion for you!
   \item[\textbf{User (U):}] Ok, tell me more?
   \item[\textbf{A:}] I would suggest to take Ben as your Manager as he is the best fit according to the brand image you want to create!
   \item[\textbf{U:}] Ok, thanks. I'll take him.
\item[]\textbf{(Task:``Banking''; \#Options = 4; Agent's Action: None)}
     \item[\textbf{U:}] I need a suggestion!
   \item[\textbf{A:}] I would suggest to take the Wallace-bank, as you have taken Ben as your manager who suits the communication style of the bank best!
    \item[\textbf{U:}] Ok, but I'll take a different bank.
\item[]\textbf{(Task:``Research''; \#Options = 4; Agent's Action: Intervention)}
   \item[\textbf{A:}] Based on your previous selections regarding your management and banking, I select autonomous driving as future research direction! \textit{Automatically makes decision and game moves on to next task...}
\end{compactitem}
\section{Simulated Proactive Dialog Environment}
Based on the data corpus, we created a simulated environment for training and evaluating the RL-based proactive DA. Therefore, we replaced the user with corpus-based user simulation and implemented an RL-based proactive DA. For adding social awareness to the agent, we developed and evaluated a supervised learning-based trust state model that allows to estimate the simulated user's current trust in the agent's action. The estimate was then included in the DA's state and its reward model.
\subsection{Simulated User}
For user simulation, the main objective was to replicate realistic user characteristics as well as task and trusting behavior for training and testing proactive dialog policies. For this purpose, the simulation relied on relevant personal and dialog data gathered from the previously described corpus collection. For creating distinct user types, the corpus' user-dependent information was used: age, gender, technical affinity, propensity to trust, domain expertise, and the Big 5 personality traits \cite{mccrae1992introduction}. Noise was added to the variables using truncated Gaussian distributions and likelihood of appearance based on corpus data distributions. Task-related behavior was simulated by generating values for the game score (task success dependent on the user's option selection), whether a user initiated a suggestion or help request, task step duration, and perceived difficulty. The probabilities for each specific user behavior were based on structured data distributions depending on the specific user type and the current dialog state. Consequently, this enabled to reproduce user behavior as a reaction to proactive system actions. For evaluating the quality of the developed user simulator, we used the Kullback-Leiber distance \cite{kullback1951information} between distributions of behavior generated by the user simulator and real users. Distances range from 0, i.e. distributions are equal, to 1, i.e distributions are completely different. The lower the distance between the distributions, the more realistic is the respective simulation. Our user simulator achieved a score of 0.172 (.142) showing realistic performance. The simulated user-specific and task-related features as well as proactive system actions were then fed as input to the trust state model which is described in the following. 
\subsection{Trust State Model}
In previous work \cite{kraus2021modelling}, we developed a novel user model for predicting trust during interaction with proactive DAs based on the collected data. As we modelled trust and its related concepts on a discrete, ordinal scale, the prediction problem was formulated as a multi-class classification task. The target classes were the distinct trust values ranging from 1 to 5. For prediction, the simulated user-specific as well as task-related features and the proactive dialog action type were quantified and concatenated into a feature vector at each system-user exchange (task step). The feature vector was then fed as input to a classifier. In previous work, we compared the performance of a Support Vector Machine (SVM), Gated Recurrent Unit (GRU) Networks, and Extreme Gradient Boosting. Typically for smaller data sets, the SVM outperformed the other approaches achieving an $F_{1}$-score of 0.533, Cohen's $\kappa$ of 0.363, a Spearman's $\rho$ of 0.426, and an extended accuracy $eA$ \cite{rach2017interaction} of 0.895 by evaluating the different algorithms using cross-validation on the data corpus. Even though the scores seem to be rather low at first sight, we deem the the proposed trust prediction model to be useful for including trust as metric for training a socially-aware agent, as trust measurement is complicated, even for human beings. This is due to trust being multi-faceted and also a latent variable that cannot be observed directly. Further, the SVM clearly outperformed the random baseline ($F_{1}$~=~0.2, $\kappa = 0.0$). Therefore, we used the trained SVM as model for predicting the user's trust state on exchange-level basis for enabling socially-aware proactive dialog modeling.
\subsection{Socially-Aware RL-Based Proactive Dialog Agent}
\label{sec:imp}
RL allows an agent to learn strategies for solving complex problems by maximising a reward, e.g. see Sutton and Barto \cite{sutton2018reinforcement} for more information. This reward is a feedback signal from the agent's environment for determining the goodness of agent behavior. To apply an RL-based approach for the adaptation of proactive dialog behavior, it was required to model the interaction between the simulated user and the agent as a Markov Decision Process (MDP). Thus, dialog states, actions, and rewards needed to be defined.

For modelling the dialog state, we included the current task step $s_{step}$ of the serious dialog game and the respective complexity $s_{complexity}$. These states represented the agent's static knowledge of the task. Dynamic knowledge was represented by integrating the last known estimated user trust value $s_{trust}$ (1-5), task success $s_{success}$ (0-10-20-30-40) and the duration $s_{duration}$ of the last task step in seconds. For modelling the action space, we relied on the taxonomy of proactive dialog acts  $a \in \{a_{none}, a_{notification}, a_{suggestion}, a_{intervention}\}$.

The reward function was modelled to promote task and also socially effective proactive behavior by the means of trustworthy interaction. As trustworthy proactive dialog behavior does not necessarily imply task effective or efficient behavior and vice versa, several aspects were taken into account for designing the reward function. To enable both trustworthy, successful, and efficient proactive dialog behavior, the reward was modelled as the sum of the rewards for estimated trust, task success, and task duration 
\begin{equation}
    r_{t} = r_{trust} + r_{success} + r_{duration} 
\end{equation}
The reward function $r_{trust}$ for each task step was modelled to reward high levels and to punish low levels of trust:
\begin{equation}
r_{trust} = \left\{
\begin{array}{ll}
\, \, \, \, 20 & \textrm{, if $s_{trust} = 5$}\\
\, \, \, \, 10 & \textrm{, if $s_{trust} = 4$} \\
\, \, \, \, 0 & \textrm{, if $s_{trust} = 3$} \\
- 10 & \textrm{, if $s_{trust} = 2$} \\
-20 & \textrm{, if $s_{trust} = 1$}
\end{array}
\right.
\end{equation}
Further, we rewarded task step success depending on the average task success score for the respective task:
\begin{equation}
r_{success} = \left\{
\begin{array}{ll}
\, \, \, \, 15 & \textrm{, if $s_{success} > mean$}\\
\, \, \, \, 10 & \textrm{, if $s_{success} = mean$} \\
\, \, \, \, 5 & \textrm{, if $s_{success} < mean$} \\
\, \, \, \, 0 & \textrm{, if $s_{success} = min$} \\
\end{array}
\right.
\end{equation}
Similarly, we rewarded task efficient behavior with regard to the duration per task step:
\begin{equation}
r_{duration} = \left\{
\begin{array}{ll}
\, \, \, \, 10 & \textrm{, if $s_{duration} \leq mean$}\\
\, \, \, \, 0 & \textrm{, if $s_{duration} > mean$} \\
\end{array}
\right.
\end{equation}
Note that we weighted the individual reward functions regarding their importance. Foremost, we aimed for trustworthy behavior. Therefore, high trust received the highest possible reward amongst all functions, whereas low trustworthy behavior even resulted in negative numerical scores. To not to neglect usability, we also rewarded successful and efficient actions. However, unsuccessful and not efficient behavior did not receive a negative reward for balancing the training more towards benefiting trustworthy dialog actions.

The proactive dialog policy was then trained in numerous interactions with simulated users. As the state space was quite large ($\approx$ 90,000 states), conventional model-free Q-leaning was not feasible. For this reason, we implemented a Deep-Q-Network (DQN) \cite{mnih2015human} approach with a stacked Multi-Layer Perceptron (MLP) for function approximation. For implementing the DQN, we utilized the stable-baseline implementation \footnote{https://stable-baselines.readthedocs.io/en/master/modules/dqn.html}. The architecture of the DQN consisted of two MLP-layers with 256 neurons, an input layer sized in the dimension of the state space, and an output player for producing the Q-values of the dialog actions. For creating the output a softmax layer was used. Using heuristic search, we applied the following hyper parameters to the DQN: For discounting future rewards, we set $\gamma =0.99$. Further, we trained the network using the RMSProp-algorithm with ADAM optimisation~\cite{kingma2014adam}, a learning rate of 0.00005, and a mini-batch size of 64. 
We used $\epsilon$-greedy for training the behavior policy with  $\epsilon$ annealed linearly from 1 to 0.1 over 15 \% of the training sample, and fixed at 0.1 thereafter. The DQN was trained on a total of 300000 training samples (task steps) or 25000 dialog games with different simulated users. For speeding up the training process, we normalized the state space using min-max scaling. The trained RL-based strategy was then evaluated against the rule-based, and the static proactive dialog strategies.
\section{Experiments and Results}
\label{sec:exp}
%
We conducted an empirical evaluation with simulated users. For comparison, we tested the RL-based proactive DA against agents using four static baseline strategies, i.e only one proactive dialog act type, e.g. None, Notification, ..., was used by the respective agent during dialog games. 

Further, we tested against a rule-based strategy that made decisions on proactive behavior depending on the user's current trust level and task complexity (see appendix for a detailed description of the rule-based strategy). 
For evaluation, we simulated 500 dialog games per strategy. For each dialog game a different user type was simulated. The set of simulated users was kept constant for each strategy to ensure comparability of the results. The amount of dialog games was selected to produce normally distributed data sets, that allowed the usage of parametric statistical significance tests. As evaluation metrics, we used the average overall trust ratings, overall task success score, and overall task duration. Significance tests for the differences between the strategies were conducted using t-tests with Bonferroni correction regarding multiple testing.

\begin{figure} 
\centering
	\includegraphics[width=\columnwidth]{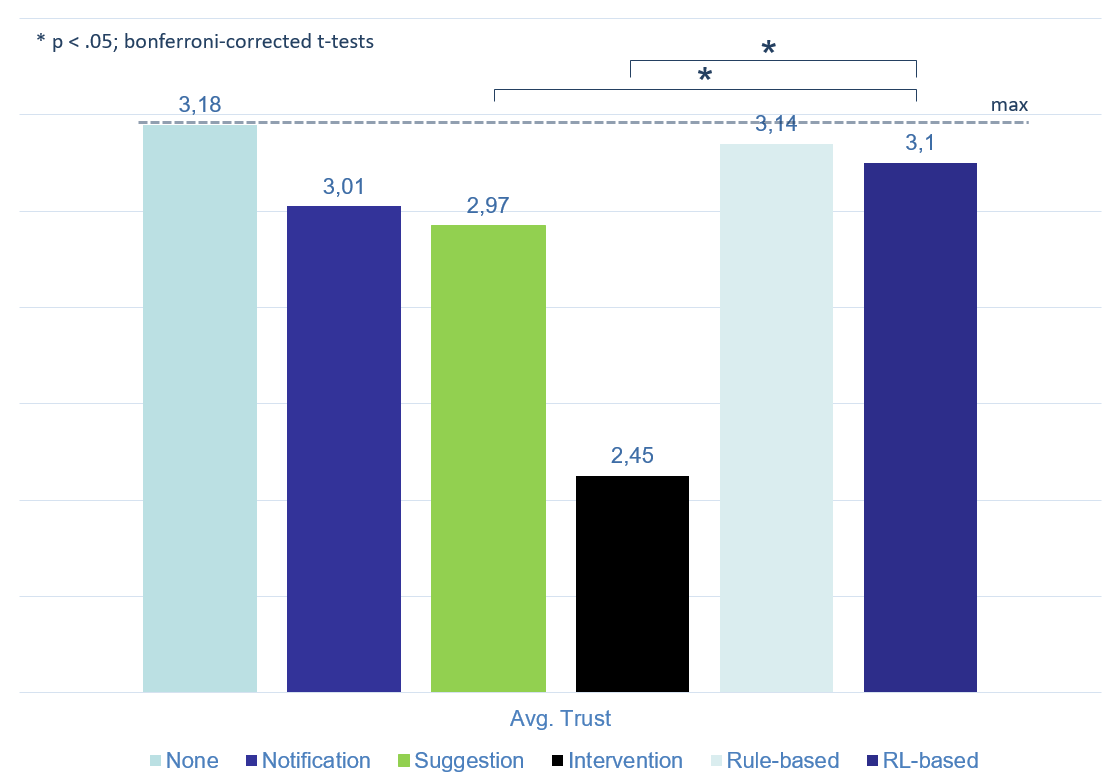}
	\caption{Average trust ratings per DA (\emph{max} indicates maximum value).}
	\label{img:rltrust}
\end{figure}
Considering the impact on HCT, the RL-based proactive DA achieved the third highest average trust value among all agents (see Fig. \ref{img:rltrust}). The differences between the RL-based agent and the agents achieving higher trust values using the \textit{None}-strategy ($p=0.315$) respectively the \textit{Rule-based}-strategy ($p=1.000$) were not significant, however. In comparison with the agents using medium-levels proactive strategies, the RL-based agent achieved significantly higher average trust values than the agent using only suggestions ($p=0.002$), but not significantly higher values than the agent using only notifications ($p=0.174$). Further, the RL-based agent scored significantly higher average trust values than the agents using the \textit{Intervention}-strategy ($p<0.001$)
\begin{figure} 
\centering
	\includegraphics[width=\columnwidth]{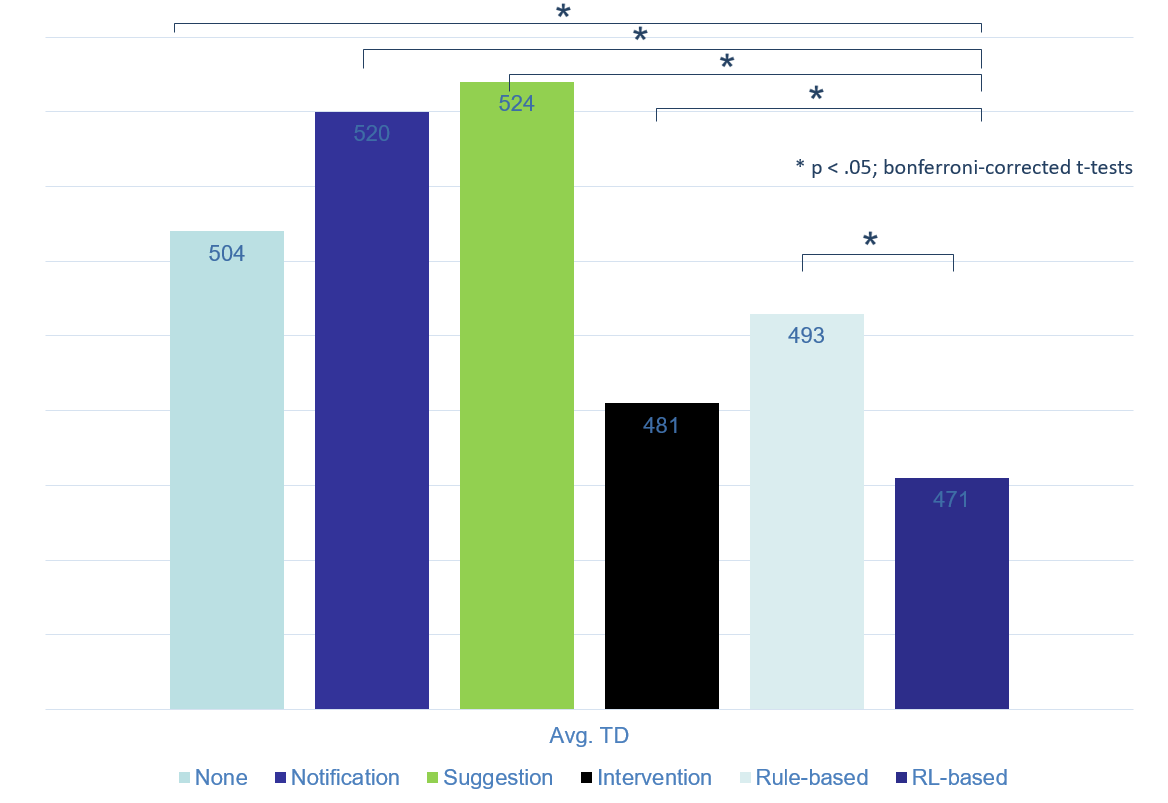}
	\caption{Average task duration per DA.}
	\label{img:rltd}
\end{figure}
Considering the impact on task duration, the agent utilising the \textit{RL-based} strategy followed by the agent applying the \textit{Intervention}-strategy led to the fastest dialog game completion (see Fig. \ref{img:rltd}). The difference between those strategies was not significant ($p=0.113$). However, the average task duration using the RL-based agent was significantly lower in comparison to all other agents (all $p<0.001$).  

\begin{figure} 
\centering
	\includegraphics[width=\columnwidth]{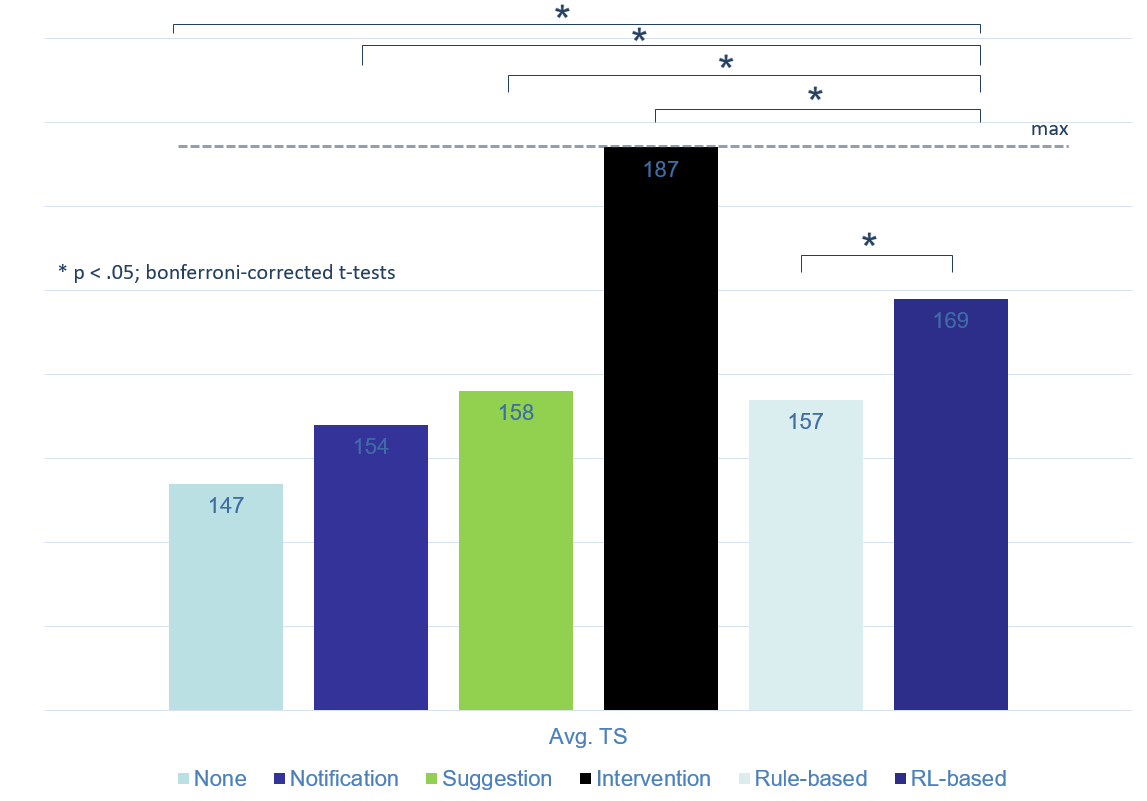}
	\caption{Average task success per DA (\emph{max} indicates maximum value).}
	\label{img:rlts}
\end{figure}

Considering the impact on task success, the RL-based proactive DA produced significantly higher task success than all other agents (all $p<0.001$) except the agent using the \textit{Intervention-strategy} (see Fig. \ref{img:rlts}). Conducting the dialog game with an agent utilising the \textit{Intervention}-strategy resulted in a significantly higher task success than interacting with the RL-based proactive DA ($p<0.001$).
\section{Discussion and Conclusion}
\label{sec:dis}
\begin{table}
    \centering
    \small
    \begin{tabular}{c|cc||c}
         \emph{Policy} & \emph{Task Efficiency} & \emph{Trust} & \emph{Cooperation}\\
         \hline
         RL-based & 0.359 & 3.10 & \textbf{1.14} \\
        None &  0.292 & \textbf{3.18} & 0.96 \\
        Notification &  0.296 & 3.01 & 0.91 \\
        Suggestion &  0.302 & 2.97 & 0.92 \\
        Intervention & \textbf{0.389} & 2.46 & 0.97 \\
        Rule-based & 0.318 & 3.14 & 1.05 \\
    \end{tabular}
    \caption{Task efficiency, trust, and cooperation scores per dialog policy.}
    \label{tab:my_label}
\end{table}
The results show that the more reactive a DA acts, the more trustworthy it is perceived. This is also in line with findings from studies with real users \cite{rau2013effects,kraus2020effects} and seems to be primarily connected with the level of control the user possesses on making the final decisions. Simultaneously, the more proactive an agent acts, the higher the task success becomes due to the agent being an expert system on the given task. Thus, a DA needs to appropriately balance reactive and different kinds of proactive behavior for enhancing cooperation. An agent needs to find the fine line on improving the user's task success, but also not to lose the user's trust which may lead to the system's disusage \cite{muir1994trust}. For making these kinds of decisions, including a degree of social awareness in the dialog policy seems to be beneficial based on our results. Observing the metrics task efficiency $\frac{success}{duration}$ and trust it becomes evident that the RL-based agent provides the best compromise between task efficient and trustworthy behavior for improving cooperation (see Table \ref{tab:my_label}). The RL-based proactive dialog policy outperforms all others except the \textit{Intervention}-strategy in task efficiency while achieving near optimal performance regarding trust which is provided by the \textit{None}-strategy. Calculating the a cooperation measure $cooperation = efficiency \cdot trust$ ultimately shows the benefit of including trust in the dialog state and/or reward function for improving cooperation (the differences between RL-based strategy and all other strategies are significant with $p < 0.05$).

Observing the proactive dialog act types the RL-based agent selected, a predominant use of the \textit{Notification}-action becomes evident (38 \% of all actions), while the extreme levels \textit{None} and \textit{Intervention} were selected well-balanced (23 \% and 25 \%). In an experiment with real users, \cite{kraus2020effects} found that notifying behavior was perceived as most appropriate with regard to a proactive system's trustworthiness. Therefore, the perception of this proactive act type maybe more invariant to the respective situation and user, while purely reactive and autonomous system behavior needs to be adapted to user and the situation.

As this study is the first of its kind utilising RL for proactive dialog management, further validation of the results are necessary. Especially, the learned policy needs to be evaluated with real users. Another limitation was the application of hand-crafted reward modelling. Here, different reward shaping methods need to be explored, which may lead to improved results. Also, other approaches need to be investigated. For example, proactive dialog policies may be learned directly from human-behavior using end-to-end approaches, e.g. by applying transformers on human proactive dialog data.
Finally, including trust is a beginning for creating socially-aware AI, but what is next? Trust is an important concept that should be considered when developing AI agents, also other important aspects need to be studied. For example, Ritschel et al. \cite{ritschel2017adapting} considered measurements of user engagement to be included in an RL-based framework for learning a robot's appropriate linguistic style. Besides, other factors of conversational intelligence, namely conscientiousness and communicability (a system's ability to self-disclose internal features and interactive principles) need to be included in sophisticated AI agents \cite{chaves2021should}.

In conclusion, we developed an RL-based proactive DA with the aim of achieving a socially and task effective interaction. For this, the agent was trained in simulated dialog environment using both trust estimate and usability measures for providing adequate proactive dialog behavior. For evaluating the approaches, we compared the RL-based proactive DA with agents using only static and rule-based proactive dialog strategies. Including trust in the dialog model for enabling adaptive dialog proved to be successful for creating socially and task effective agents. Our approach achieved the best compromise of contributing to task efficiency, but also acting in a trustworthy manner. In future work, we plan to validate the results in a study with real users.

\bibliographystyle{IEEEtran}

\bibliography{template.bib}
\newpage
\appendix
\section{Rule-based Proactive Dialog Strategy}
\begin{figure}[h!] 
\centering
	\includegraphics[width=\textwidth]{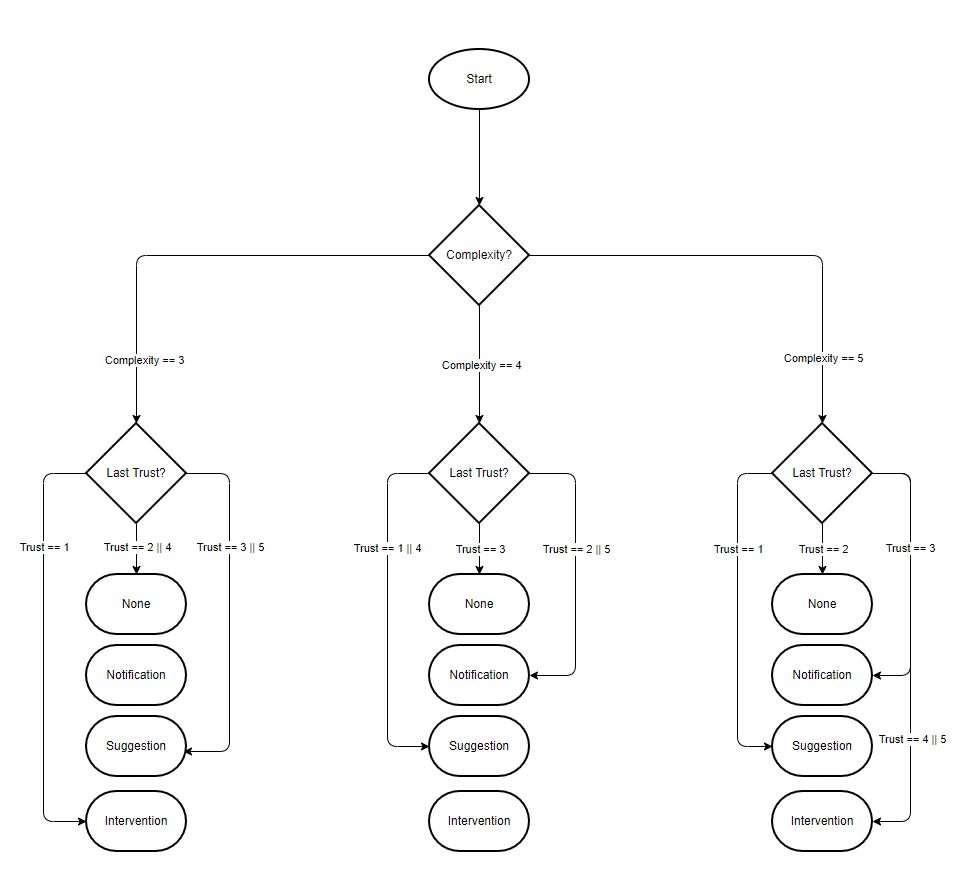}
	\label{img:rulechart}
\end{figure}


\end{document}